\documentclass[letterpaper, 10 pt, conference]{ieeeconf}  

\IEEEoverridecommandlockouts                              

\usepackage{epsfig} 

\let\labelindent\relax

\usepackage{times}
\usepackage{graphicx}
\usepackage{amsmath}
\usepackage{amsthm}
\usepackage{amssymb}
\usepackage{amsopn}
\usepackage{amstext}
\usepackage{amsfonts}
\usepackage{cancel}
\usepackage[space]{cite}
\usepackage{soul}
\usepackage{cuted}
\usepackage{lipsum,multicol}
\usepackage{stfloats}
\everymath{\displaystyle}
\usepackage{booktabs}
\usepackage{xfrac}
\usepackage{balance}
\usepackage{color}
\usepackage{mathtools}
\usepackage{tabularx}
\newcolumntype{L}{>{\raggedright\arraybackslash}X}
\newcolumntype{C}{>{\centering\arraybackslash}X}
\newcolumntype{R}{>{\raggedleft\arraybackslash}X}
\usepackage{bm}
\usepackage{diagbox}
\usepackage{float}
\usepackage{epstopdf}
\usepackage{url}
\usepackage{multirow}
\usepackage{tikz}
\usepackage[linkcolor=black,citecolor=black,urlcolor=black,colorlinks=true]{hyperref}
\usepackage{enumitem}
\usepackage[ruled,vlined,linesnumbered]{algorithm2e}
\usepackage{threeparttable}

\DeclareMathAlphabet\mathbfcal{OMS}{cmsy}{b}{n}

\title{\LARGE \bf
AS-LIO: Spatial Overlap Guided Adaptive Sliding Window LiDAR-Inertial Odometry for Aggressive FOV Variation
}

\author{Tianxiang Zhang$^{1}$  ~ Xuanxuan Zhang$^{1}$ ~ Zongbo Liao$^{1}$ ~ Xin Xia$^{2}$ ~ You Li$^{1}$ %
\thanks{
This work was supported in part by the National Key R\&D Program of China (2022YFE0139300, 2022YFB3903800), the Major Science and Technologu Projects in Hubei Province (2022AAA009), and the National Natural Science Foundation of China (42274052).
}%
\thanks{$^{1}$Tianxiang Zhang, Xuanxuan Zhang, Zongbo Liao, and You Li are with State Key Laboratory of Information Engineering in Surveying, Mapping and Remote Sensing (LIESMARS), Wuhan University, 430072, China. You Li is the corresponding author. Email: 
\tt \footnotesize \{\small cyberkona; xuanxuanzhang; liaozb; liyou\footnotesize \}\small @whu.edu.cn }%
\thanks{$^{2}$Xin Xia is with Department of Civil and Environmental Engineering, UCLA Samueli School of Engineering, Los Angeles, CA 90095. Email: \tt \small x35xia@ucla.edu}%
} 

\begin{document}
\maketitle
\thispagestyle{empty}
\pagestyle{empty}

\begin{abstract}

LiDAR-Inertial Odometry (LIO) demonstrates outstanding accuracy and stability in general low-speed and smooth motion scenarios. However, in high-speed and intense motion scenarios, such as sharp turns, two primary challenges arise: firstly, due to the limitations of IMU frequency, the error in estimating significantly non-linear motion states escalates; secondly, drastic changes in the Field of View (FOV) may diminish the spatial overlap between LiDAR frame and pointcloud map (or between frames), leading to insufficient data association and constraint degradation.

To address these issues, we propose a novel Adaptive Sliding window LIO framework (AS-LIO) guided by the Spatial Overlap Degree (SOD). Initially, we assess the SOD between the LiDAR frames and the registered map, directly evaluating the adverse impact of current FOV variation on pointcloud alignment. Subsequently, we design an adaptive sliding window to manage the continuous LiDAR stream and control state updates, dynamically adjusting the update step according to the SOD. This strategy enables our odometry to adaptively adopt higher update frequency to precisely characterize trajectory during aggressive FOV variation, thus effectively reducing the non-linear error in positioning. Meanwhile, the historical constraints within the sliding window reinforce the frame-to-map data association, ensuring the robustness of state estimation. Experiments show that our AS-LIO framework can quickly perceive and respond to challenging FOV change, outperforming other state-of-the-art LIO frameworks in terms of accuracy and robustness.

\end{abstract}


\section{INTRODUCTION}

Owing to the ability to endow robots with the perception of the position and surroundings, Simultaneous Localization and Mapping (SLAM) technology has become fundamental for implementing more complex robotic applications~\cite{cadena2016past}. 
The LiDAR-Inertial Odometry (LIO) stands out for its superior stability~\cite{zou2021comparative,lee2024lidar} and has been extensively applied in fields such as autonomous robotics and self-driving vehicles~\cite{roriz2021automotive,ding2020lidar,yoo2018mems}. Light Detection and Ranging (LiDAR) can offer dense and direct distance measurement~\cite{wang2020mems}, avoiding the issue of scale inconsistency~\cite{strasdat2010scale} inherent in visual-based odometry. Compared to cameras limited by narrow dynamic ranges, LiDAR is insensitive to environmental lighting conditions~\cite{li2021towards}, ensuring operational reliability across day and night scenarios. 
Additionally, LiDAR-based odometry can achieve registration directly on dense pointcloud frames without feature extraction~\cite{xu2022fast}, while vision-based odometry may risk failures in optical flow tracking or visual feature extraction~\cite{saputra2018visual,debeunne2020review,yuan2023sdv}, leading to severe divergence in state estimation.

\begin{figure}[t]
    \begin{center}
        {\includegraphics[width=1\columnwidth]{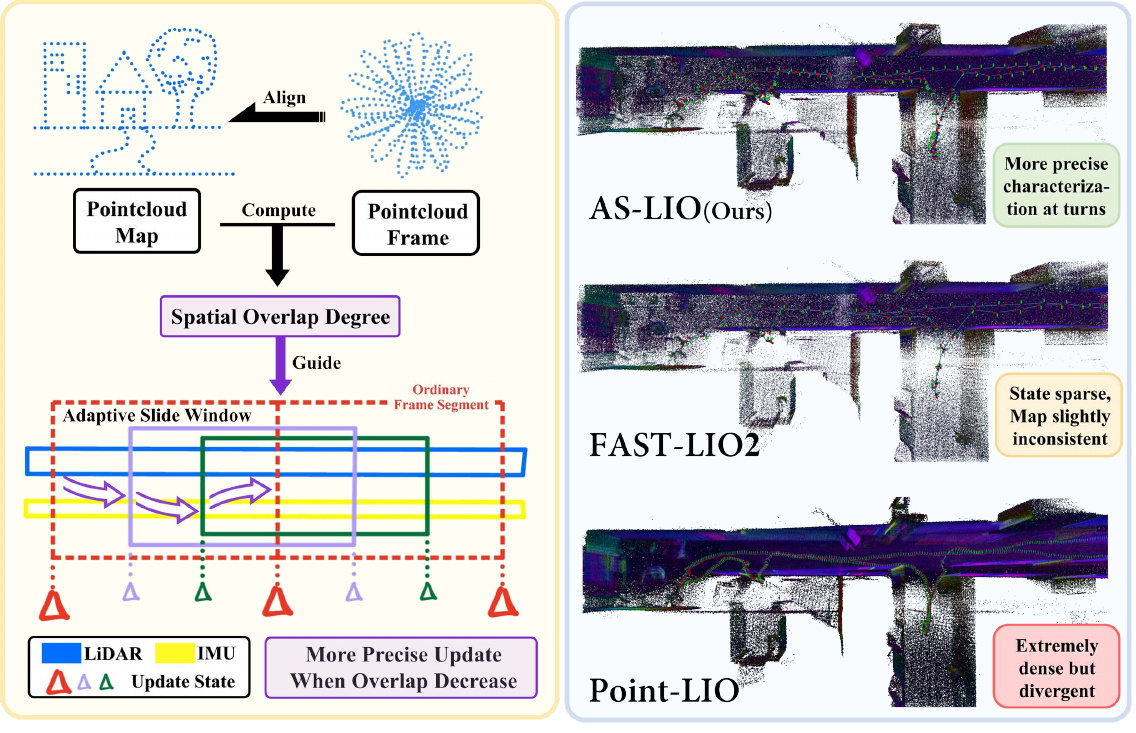}}
    \end{center}
    \vspace{-0.3cm}
    \caption{\label{fig:intro_part}Our AS-LIO framework (Left) and result comparision (Right). We utilize the frame-to-map spatial overlap degree to assess the impact of FOV change on pointcloud alignment, then accordingly adjust the update step of the sliding window and maintain the history constraints.}
    \vspace{-0.4cm}
\end{figure}

Existing LIO frameworks typically adopt a strategy based on discrete frame registration~\cite{bai2022faster,qin2020lins,yuan2023semi,reinke2022locus}. This approach involves segmenting discrete frames from continuous pointcloud stream at a predetermined temporal scale. Subsequently, observations from the Inertial Measurement Unit (IMU) are utilized to compensate for partial motion distortion~\cite{ye2019tightly,chen2023direct} and provide prior state constraints. Then the pointcloud registration can be performed based on the rigid body assumption. 
However, when confronted with significant change in FOV, like sharp turn scenarios, this strategy faces a dilemma in selecting the appropriate temporal scale for pointcloud frames. 

On one hand, expanding the temporal scale will cause an accumulation of IMU noise during forward propagation (i.e., IMU integration), thus exacerbating the prior error and the residual motion distortion within frames. Concurrently, aggressive FOV change often coincides with rapid alteration in orientation, which implies an increase in the approximate linearization error of the propagation and the measurement matrices~\cite{xu2022robots}. These factors could collectively restrict the state estimation from further convergence~\cite{yang2020graduated}. 
On the other hand, shortening the temporal scale will correspondingly raise the odometry frequency. Characterizing a violent motion process through more dense states could mitigate the motion distortion and the system non-linear error  to some extent. However, due to the reduction of LiDAR points within the frame, the spatial overlap from current frame to pointcloud map (or neighboring frames) reduced, potentially resulting in insufficient data association or even degenerate constraints, lowering the robustness of the odometry system.

Moreover, while using LiDAR with a larger FOV can mitigate the impact of FOV changes caused by platform movement, it cannot prevent FOV changes resulting from external environmental occlusions. Semi-solid LiDAR with a limited FOV may experience aggressive FOV changes due to its own movement at sharp turns. In contrast, spinning LiDAR with a 360-degree FOV is more likely to encounter aggressive FOV changes when rapidly passing through doorways, caverns, bridge tunnels, or similar scenarios with occlusions.

To achieve a balance between accuracy and robustness, we propose a novel adaptive sliding window framework guided by the spatial overlap degree (SOD), specifically designed to address the challenging scenarios with aggressive FOV variation. 
We propose a new metric to evaluate the SOD from the LiDAR frames to the pointcloud map in real-time. Unlike the angular velocity measurement from the IMU, which indirectly indicates the degree of FOV change, the SOD we proposed can provide a more direct evaluation of the impact of the immediate FOV change on pointcloud registration. 
Upon this metric, we develop a novel adaptive sliding window framework for LIO. The sliding window manages the continuous LiDAR stream and controls the state update. The pointcloud inside the window is denoted as a frame. The update step is dynamically adjusted according to the SOD, and the step size actually corresponds to the shift time of the sliding window in our framework.

For example, the shift time will be shortened when SOD decrease in order to increase the update frequency and delineate the trajectory with more states, hence suppressing nonlinear errors and motion distortions. Additionally, the historical pointcloud within the sliding window can serve as supplementary constraints, compensating the frame-to-map overlap degree and diminishing the risk of registration degradation. As a result, while the positioning error of LIO can achieve further convergence, the stability of the state estimation can be guaranteed as well. 

In summary, our main contributions are as follows:

\begin{itemize}

\item We introduce a new metric to assess the spatial overlap degree from the LiDAR frames to the pointcloud map, and utilize a 3D soft-margin voxel map to achieve real-time and stable computation. 
In contrast to IMU measurement, the spatial overlap degree we proposed can more directly represent the adverse impact of aggressive FOV variation on pointcloud registration.
 
\item We implement a novel adaptive sliding window framework for LIO, managing the continuous LiDAR stream and dynamically adjusting the shift time of sliding window based on the spatial overlap degree mentioned above. 
Our framework not only facilitates further convergence of the positioning errors via the adjustment of shift time and hence update frequency, but also ensures the robustness of the state estimation with the historical pointcloud constraints.

\item We conduct extensive experiments on AS-LIO in a series of indoor and outdoor real-world datasets. 
The experimental results demonstrate that our proposed AS-LIO is more accurate and robust compared to the state-of-the-art open-source methods.

\end{itemize}

\section{RELATED WORK}

Contemporary LiDAR(-inertial) odometry frameworks typically employ a separate-frame strategy for registration. 
During the scanning process, the continuous motion of LiDAR can result in inconsistencies in coordinate systems of the laser points collected at various instances, thereby inducing motion distortion phenomena within pointcloud frames~\cite{vizzo2023kiss}. 
Moreover, such motion distortion can impede the convergence of residuals during the registration phase. 
In scenarios of intense motion, significant distortions may further exacerbate this issue, leading to severe divergence in state estimation.

LOAM~\cite{zhang2014loam} is the most classical LiDAR odometry framework, which extracts edge and plane features from pointcloud for registration and estimates the end state of LiDAR frames. 
However, it relies on a constant velocity model to compensate for distortion, and this introduces significant nonlinear errors. CT-ICP~\cite{dellenbach2022ct} takes into account that the error in the end state of previous frame may restrain the convergence in current frame via motion distortion. 
Therefore, it performs joint optimization of the beginning and end states of LiDAR frames and compensates for distortion with interstate interpolation. 
However, the registration results are sensitive to the logical constraint between the beginning state of the current frame and the end state of the previous frame.

Frameworks such as FAST-LIO~\cite{xu2021fast,xu2022fast} and LIO-SAM~\cite{shan2020lio} utilize the IMU integration to deal with the distortion. FAST-LIO implements a backward propagation to compensate for motion distortion and conducts state estimation by Error State Iterative Kalman Filter. 
But limited IMU frequency will amplify the residual motion distortion in violent motion, and thus degrade the odometry's performance.
LIO-SAM is the first factor graph LIO framework and employs IMU pre-integration to de-skew pointclouds. It performs alignment between neighboring frames with geometric features and then selects key frames for optimization.
Nonetheless, the long period of integration will increase cumulative errors, and the heavy computational cost makes it difficult to maintain stable real-time performance. 

To address the issue of increased motion distortion and linearization approximation errors under intense motion, many frameworks have adopted variable frame length strategies.
SR-LIO~\cite{yuan2022sr} segments and reconstructs the original LiDAR sweep inputs, thereby increasing the frequency of state updates. However, its reconstruction splits the original frame into two fixed sub-frames and then concatenates them head-to-tail, without considering the distortion correction of the previous sub-frame after the update of the middle state of the original frame. FR-LIO~\cite{liu2023fr} proposes a method to segment original frames based on IMU observations and utilizes the constraints of neighboring subframes to smooth estimation results to cope with the potential degradation risk of subframe registration. However, this post-processing approach significantly increases the additional computational overhead. But this post-processing approach greatly increases the additional computation overhead as the subframe states increasing.

Furthermore, Point-LIO~\cite{he2023point} adopts a point-by-point strategy to avoid the motion distortion issue, achieving state estimation under extreme aggressive motion with higher frequency updates.
However, the approach of single-point observation forsakes the statistical advantages of dense pointcloud, making it more sensitive to dynamic objects and outliers, and thus adversely affect the stability of state estimation.

In contrast, our proposed AS-LIO framework can perceive aggressive FOV variation based on the spatial overlap degree between LiDAR frames and the map, and then we utilize an adaptive sliding window to accommodate such drastic changes.
This approach serves a dual purpose: it not only suppresses motion distortion and nonlinear errors by controlling the update step and frequency, but also ensures the stability of state estimation through the maintenance of historical point cloud constraints.

By integrating these strategies, our AS-LIO framework effectively addresses the previously mentioned issues of motion distortion exacerbation and the challenges in maintaining accurate state estimation under intense motion scenarios. 
Additionally, by achieving the balance between the solution complexity and the solution frequency, our method mitigates the overall computational burdens while performing fine-grained characterization of the intense motion trajectories.
This holistic approach not only enhances the precision and reliability of odometry in challenging scenarios but also offers a new benchmark for LiDAR-Inertial Odometry by providing a novel solution to the inherent limitations of previous methodologies.

\section{METHODOLOGY}

\subsection{System Overview}

\begin{figure}[h]
    \begin{center}
        {\includegraphics[width=1\columnwidth]{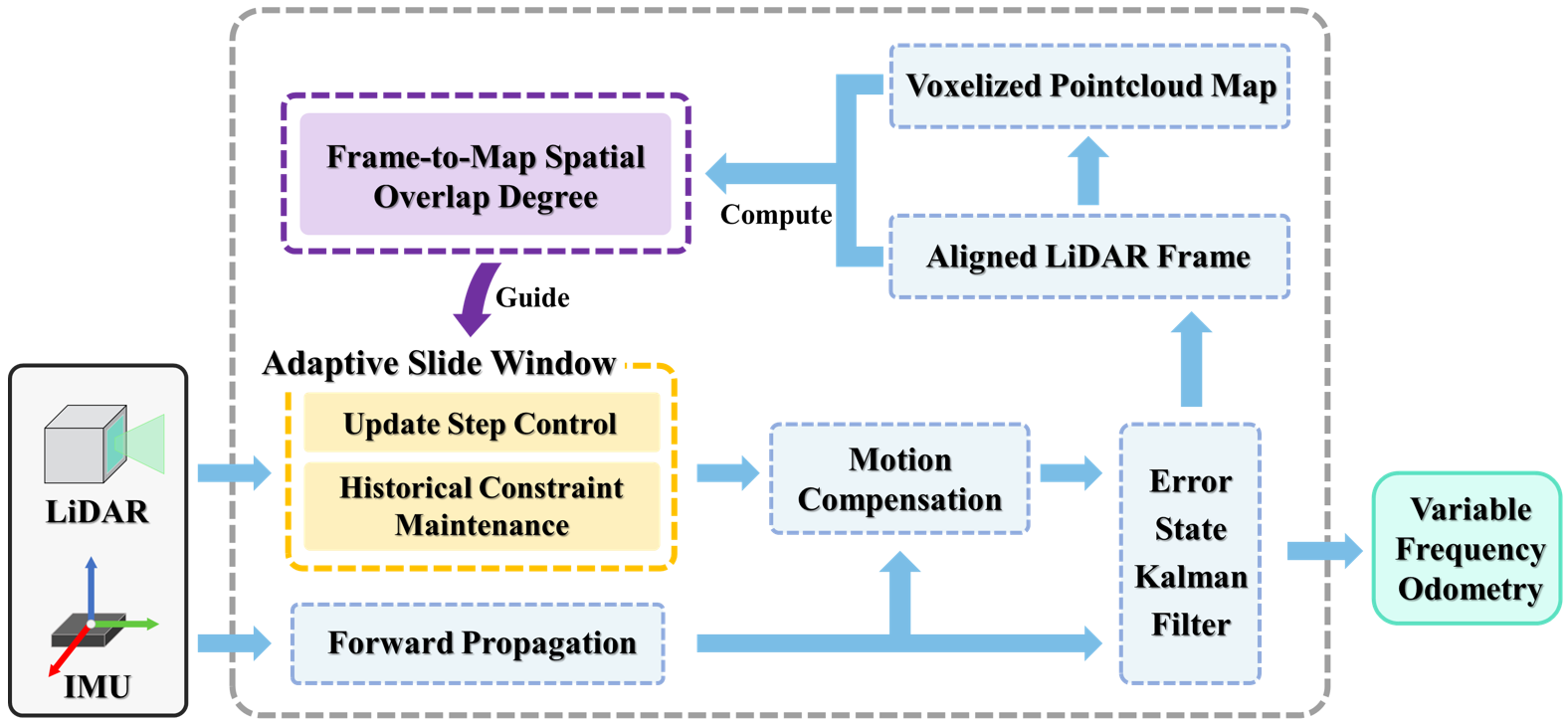}}
    \end{center}
    \vspace{-0.2cm}
    \caption{\label{fig:System Overview}The system overview of our AS-LIO framework.}
    \vspace{-0cm}
\end{figure}

The overview of our AS-LIO framework is shown in Fig.~\ref{fig:System Overview}.
The continuous LiDAR input stream is fed into the adaptive sliding window module to extract the pointcloud frame containing the latest and historical observations based on the update step. 
Then, the pointcloud frame is compensated by IMU and fed into the Error State Kalman Filter (ESKF) state estimation to obtain the dynamic frequency odometry output (from 10 Hz to 125 Hz).
The aligned LiDAR frame is first projected to the 3D voxelized pointcloud map previously accumulated, to compute the frame-to-map SOD, after that merging with the pointcloud map.
The SOD, reflecting the impact of FOV change on pointcloud alignment, is then used to guide the update step control in the sliding window for the next state update.

\subsection{Spatial Overlap Degree Calculation}

In the real world, when scanning the external environment using sensors (e.g., LiDAR, mmWave Radar, etc.), it is often impossible to capture the exact same point positions. 
For 3D pointclouds with floating-point coordinates, even if we align several frames of the same scenes with their true poses, there will rarely occur points with identical coordinates.
Furthermore, in pointcloud registration and related algorithms, the estimation of pose primarily relies on the shared geometric representation of scenes across pointclouds, rather than a set of points with exactly the same coordinates.

Therefore, for the two aligned pointcloud sets ${P_s}$ and ${P_t}$, and the points ${p_s}$ and ${p_t}$ within them ($p_s \in P_s$ and $p_t \in P_t$), if the spatial distance between ${p_s}$ and ${p_t}$ is sufficiently small ($\left| \left| p_s - p_t \right| \right|_2 < \epsilon$), then it can be assumed that the couple of points are located in the overlapping region of ${P_s}$ and ${P_t}$. For ease of computation, we employ the 3D voxel space to represent the accumulated pointcloud map,
and define the point set in common occupied space as the overlapping part of aligned pointcloud frames.

\begin{equation}
    \label{equation1}
    \begin{gathered}
        \Omega_{s} \doteq \mathit{3d\_voxel\_hash} (\mathit{P}_{s})\\
        \Omega_{t} \doteq \mathit{3d\_voxel\_hash} (\mathit{P}_{t}) \\
        \Omega_{{s \cdot t}} = \Omega_{s} \cap \Omega_{t} \\
        \mathit{Q}_s^t = \mathit{P}_{s} \cap \Omega_{{s \cdot t}}\\
        \mathit{Q}_t^s = \mathit{P}_{t} \cap \Omega_{{s \cdot t}} 
    \end{gathered}
\end{equation}

We utilize a three-layer embedded hash table to record the 3D voxels hit by the pointcloud, and the size of 3D voxels is pre-defined.
By employing the $3d\_voxel\_hash(\cdot)$ function in Eq.~\ref{equation1}, we obtain the 3D voxel spaces $\Omega_s$ and $\Omega_t$ occupied by $P_s$ and $P_t$ respectively. 
Then we determine the intersection space $\Omega_{s \cdot t}$ between $\Omega_s$ and $\Omega_t$ (i.e., the shared occupied voxel space of $P_s$ and $P_t$). Next, we find the point sets $Q_s^t$ and $Q_t^s$ that reside in the overlapping space $\Omega_{s \cdot t}$ for $P_s$ and $P_t$ ($Q_s^t \in P_s | Q_s^t \in \Omega_{s \cdot t}$ and $Q_t^s \in P_t | Q_t^s \in \Omega_{s \cdot t}$). Denote $Num(\cdot)$ as the point number in the point set. Naturally, we derive the overlap score between pointcloud frames as

\begin{equation}
    \label{equation2}
    \begin{gathered}
        {O}_s^t = \mathit{Num} (\mathit{Q}_s^t) /  \mathit{Num} (\mathit{P}_s)\\
        {O}_t^s = \mathit{Num} (\mathit{Q}_t^s) /  \mathit{Num} (\mathit{P}_t)
    \end{gathered}
\end{equation}

Due to variation in the element number and density across pointcloud frames, the relative scores of spatial overlap between frames also differ. For convenience, we define $O_s^t$ represents the overlap score of $P_s$ relative to $P_t$, while $O_t^s$ represents the score of $P_t$ relative to $P_s$. Therefore, for the pointcloud map $M$ accumulated from LIO and the pointcloud frame $f$ under the posterior pose, the overlap score of $f$ to $M$ should be

\begin{equation}
    \label{equation3}
    \begin{gathered}
        {O}_f^M = \mathit{Num} ({Q}_f^M) /  \mathit{Num} (f)
    \end{gathered}
\end{equation}

Given the discrepancy between the posterior pose and the true pose, we adopt a soft-margin strategy to enhance the stability of the frame-to-map spatial overlap computation. This involves appropriately extending the 3d voxel space occupied by the pointcloud map $M$ at the unit scale of voxels, i.e., expanding the $\Omega_M$. Denote the extended space as $\Omega_{M^*}$, the extended part as $\Omega_{M^i}$, with $i$ denoting the shortest unit distance from the extended voxel to the occupied voxel.
We set an upper limit $d$ on $i$ to adjust the extension distance of the 3D voxel space ($i < d$ and $i \in \mathbb{N}$). 

\begin{equation}
    \label{equation4}
    \begin{gathered}
        \Omega_{M^*} = \Omega_{M} + \sum_{i=1}^{d} \Omega_{M^i}
    \end{gathered}
\end{equation}

Then, we assign the points from $f$ and in $\Omega_{M^*}$ with decreasing weights $\beta_i$ as the extension distance $i$ increases. (For $\beta_i$ the descent gradient can be set to a reasonable value, and for $d$ we simply set $d=3$ in our test samples.) Consequently, the modified overlap score $O_f^{M^*}$ should be

\begin{equation}
    \label{equation5}
    \begin{gathered}
        {O}_f^{M^*} = 
        \begin{bmatrix}
        1 & \beta_1 & ... & \beta_d
        \end{bmatrix}
        \begin{bmatrix}
        \mathit{Num} ({Q}_f^{M}) \\
        \mathit{Num} ({Q}_f^{M^{1}}) \\
        ... \\
        \mathit{Num} ({Q}_f^{M^{d}})
        \end{bmatrix}
        /  \mathit{Num} (f)
    \end{gathered}
\end{equation}

Noting that $\beta_0 = 1$ and $M = M_0$, the simplified overlap score can be expressed as

\begin{equation}
    \label{equation6}
    \begin{gathered}
        {O}_f^{M^*} = 
        \sum_{i=0}^{d} \beta_i *  \mathit{Num} ({Q}_f^{M_i})
        /  \mathit{Num} (f) * 100 \%
    \end{gathered}
\end{equation}

Based on the 3d soft-margin voxel map, we can quickly and accurately compute the frame-to-map SOD. 
Since the voxel space occupied by pointcloud map is appropriately expanded, it will lead to similar SOD results at small pose differences. Therefore, when the posterior pose only differs slightly from the true pose, the soft-margin strategy ensures the computed SOD is highly approximated to the SOD under the true pose. Additionally, it also ensures the SOD changes remain continuous and responsive during platform movement.
Subsequent experiment sections show that the SOD has a higher signal-to-noise ratio compared to the IMU data. For turning scenes with a vast FOV and a low speed, our overlap degree can more significantly detect when the FOV change weakens the pointcloud constraints.

\subsection{Adaptive Sliding Window Management}

Conventional LIO frameworks typically segment the continuous LiDAR stream to obtain discrete pointcloud frames for registration. The errors within the pointcloud frames primarily stem from individual laser point measurement and general motion distortion. 
For LIO frameworks, the motion distortion within frames can be compensated through the motion observations of the IMU to some extent.
Therefore, for the residual motion distortion after compensation, it is mainly influenced by the IMU frequency and accuracy, as well as the predefined temporal scale of LiDAR frames.

In scenarios with significant change in LiDAR FOV, controlling the error in LIO systems confronts with a dilemma. Shortening the temporal scale of LiDAR frames can effectively lower the intra-frame distortion and the nonlinearity in state estimation. However, excessively short LiDAR frames may lead to insufficient constraints and registration degradation, thereby impacting the system robustness.

To enhance the positioning accuracy while ensuring the system robustness as possible, we propose a strategy for these challenging scenarios with aggressive FOV variation.
This strategy involves reducing the temporal scale of the latest LiDAR frame based on a specific metric to suppress positioning errors, and incorporating historical constraint information to ensure registration stability. Consequently, we implement an adaptive sliding window framework for continuous LiDAR stream processing.

Based on the spatial overlap metric $O_f^{M^*}$ proposed in the last section, we dynamically adjust the update step length $shift\_time$ of the sliding window. 
When the overlap score between the previous frame under posterior pose and the map is high enough, it can be inferred that the FOV change are relatively smooth during current movement, and no additional operations are required. 
When the overlap score is too low, it indicates that the LIO system is facing severe FOV change. In this case, the $shift\_time$ of the sliding window should be shortened accordingly, reducing the temporal scale of the latest pointcloud observations involved in the registration. The historical observations within the sliding window serve as auxiliary constraints, with appropriate weight adjustment and filtering, to participate in the state update along with the latest observations. Denote $\lceil \cdot \rceil$ as an upward rounding integer, then we have

\begin{equation}
    \label{equation7}
    \begin{gathered}
    \mathit{seg\_time} = \lceil (1 - O_f^{M^*}) / \mathit{seg\_step} \rceil + 1 \\
    \mathit{shift\_time} = \mathit{frame\_length} * 2 / \mathit{seg\_time}
    \end{gathered}
\end{equation}

For simplicity, we set the total temporal scale of the sliding window to a fixed length ${frame\_length}$, and define the decline gradient of the SOD as $seg\_step$. 
Based on Eq.~\ref{equation7}, we calculate the intermediate variable $seg\_time$, which represents the slicing times of $frame\_length$ and controls the reduction of the $shift\_time$, and then derive the update step $shift\_time$ of the sliding window.
As the sliding window paces, the ratio of historical observations to the latest within the window is $(frame\_length - shift\_time)/shift\_time$. 
When ${frame\_length}$ remains constant but $shift\_time$ is shortened, the current sliding window will utilize fewer latest pointcloud observations while retaining more historical observations for the next state update.
Moreover, it is evident that the parameter $seg\_step$ determines the sensitivity of the $seg\_step$ to the $O_f^{M^*}$ decline.

\begin{equation}
    \label{equation8}
    \begin{gathered}
    \mathit{echo\_time} = (\mathit{seg\_time} \leq 2) \; \mathit{?} \; 1 : \mathit{seg\_time}
    \end{gathered}
\end{equation}

In consideration of the continuity of the motion process, we introduce an echo strategy to refine our framework.
Derived from Eq. \ref{equation8}, the $echo\_time$ represents the updated times that the sliding window steps according to $shift\_time$ as expected, where the $(\cdot)\mathit{?}(\cdot)$:$(\cdot)$ refers to the ternary conditional operator in {C}/{C++} language.
If the overlap degree $O_f^{M^*}$ further decreases in subsequent updates, leading to a reduction in $seg\_time$, then the $shift\_time$ and $echo\_time$ wil be recalculated for more dense state refinement.

\subsection{ESKF State Estimation}

The AS-LIO we proposed employs the classical ESKF for state estimation~\cite{sola2012quaternion} . Based on the kinematic model and IMU measurement model, the state $\mathbf{x}$ , input $\mathbf{i}$, and noise $\mathbf{n}$ can be defined as follows:
\begin{equation}
    \label{equation9}
    \begin{split}
        \mathbf x & \doteq \begin{bmatrix} \mathbf{R}_I^G & \mathbf{p}_I^G & \mathbf{v}_I&\mathbf{b}_{\bm \omega} & \mathbf{b}_{\mathbf a} & \mathbf{g}^G \end{bmatrix}^T \\
        \mathbf i & \doteq \begin{bmatrix} {\bm \omega}_m & {\mathbf a}_m\end{bmatrix}^T \\
        \mathbf n & \doteq \begin{bmatrix}\mathbf n_{\bm \omega}&\mathbf n_{\mathbf a}&\mathbf n_{\mathbf b\bm \omega}&\mathbf n_{\mathbf b\mathbf a}\end{bmatrix}^T 
    \end{split}
\end{equation}

Where $\mathit{G}$ and $\mathit{I}$ denote the global and the IMU frames, $\mathbf{R}_I^G$, $\mathbf{p}_I^G$ and $\mathbf{v}_I^G$ are the orientation, position and linear velocity of IMU in global frame $\mathit{G}$, $\mathbf{g}^G$ is the local gravity. 
The $\mathbf{n}_{\bm \omega}$, $\mathbf{n}_{\mathbf a}$ are the noises of angular velocity ${\bm \omega}_m$ and linear acceleration ${\mathbf a}_m$, $\mathbf{b}_{\bm \omega}$, $\mathbf{b}_{\mathbf a}$ are the IMU bias with $\mathbf {n}_{\mathbf{b}\bm \omega}$ and $\mathbf n_{\mathbf b\mathbf a}$ as their own noises in Gaussian-Markov model.

Furthermore, the state $\mathbf{x}$ can be considered as a high-dimensional manifold locally homeomorphic to Euclidean space $\mathbb{R}$. The operations $\boxplus$ and $\boxminus$ on the manifold can be defined according to~\cite{hertzberg2013integrating}. 
Denote the error state as $\delta \mathbf{x}$, we can derive the recursion for $\delta \mathbf{x}$ and its covariance $\mathbf{P}$ from the propagation.
\begin{equation}
    \label{equation10}
    \begin{gathered}
        \quad \quad \delta \mathbf{x}_i  \doteq \begin{bmatrix} \delta {\bm \theta} & \delta \mathbf{p}_I^G & \delta \mathbf{v}_I& \delta \mathbf{b}_{\bm \omega} & \delta \mathbf{b}_{\mathbf a} & \delta \mathbf{g}^G \end{bmatrix}^T \\
        \quad \delta \hat{\mathbf{x}}_{i+1} = \mathbf{F_a} \delta \check{\mathbf{x}}_{i} + \mathbf{F_n} \mathbf{n}_i \\
        \quad \hat{\mathbf{P}}_{i+1} = \mathbf{F_a} \check{\mathbf{P}}_{i} \mathbf{F_a}^T + \mathbf{F_n} \mathbf{Q} \mathbf{F_n}^T
    \end{gathered}
\end{equation}

Assuming the extrinsic $\mathbf{T}_{L}^{I}$ between LiDAR and IMU is known, the pointcloud can be transformed to frame $\mathit{I}$ under the prior pose $\hat{\mathbf{x}}_i$.
For a point $\mathbf{p}$ in LiDAR frame, its neighboring cluster can be searched by KNN in the map. A plane is then fitted to this cluster to obtain the normal vector $\mathbf{u}$. 
Denote one point of the cluster as $\mathbf{q}$ and $( \cdot )_{\wedge} $ as the upper triangular matrix of vector, the residual $\mathbf{z}$ and measurement matrix $\mathbf{H}$ are as follows:
\begin{equation}
    \label{equation11}
    \begin{gathered}
        \mathbf{z} = \mathbf{u}(\hat{\mathbf{p}}^G - \mathbf{q}^G)^T \\
        \mathbf{H} = \begin{bmatrix} -\mathbf{u}\hat{\mathbf{R}}_I^G(\mathbf{T}_{L}^{I} \mathbf{p}^L)_{\wedge} & \mathbf{u} & \mathbf{0}_{3 \times 12} \end{bmatrix}
    \end{gathered}
\end{equation}

Considering the nonlinear errors in the $\mathbf{F_a}$, $\mathbf{F_n}$ and $\mathbf{H}$, the estimation may require iteration to promote convergence. Let $\mathbf{A}$ denote the Jacobian of update component with respect to $\delta \mathbf{x}$ , according to~\cite{xu2021fast}, the update equation is:
\begin{equation}
    \label{equation12}
    \begin{gathered}
        \mathbf{K} = \mathbf{P}\mathbf{H}^T(\mathbf{H}\mathbf{P}\mathbf{H}^T + \mathbf{R})^{-1} \\
        \hat{\mathbf{x}}_{i+1}^{j+1} = \hat{\mathbf{x}}_{i+1}^{j} \boxplus [ -\mathbf{K}\mathbf{z} - (\mathbf{I} - \mathbf{K}\mathbf{H}) (\mathbf{A}^j)^{-1} (\hat{\mathbf{x}}_{i+1}^{j} \boxminus \hat{\mathbf{x}}_{i+1} ) ]
    \end{gathered}
\end{equation}

When the state update is less than a threshold, the solution can be assumed to converge.
\begin{equation}
    \label{equation13}
    \begin{gathered}
        \check{\mathbf{x}}_{i+1} \Leftarrow \hat{\mathbf{x}}_{i+1}^{j+1}, \; \check{\mathbf{P}}_{i+1} \Leftarrow ( \mathbf{I} -  \mathbf{K}\mathbf{H} )\hat{\mathbf{P}}_{i+1}
    \end{gathered}
\end{equation}

With the concise and effective of ESKF, our AS-LIO can achieve stable and reliable online state estimation, the test results are shown in the next experimental section.

\section{EXPERIMENT RESULTS}

\begin{figure}[t]
    \vspace{1.5mm}
    \begin{center}
        {\includegraphics[width=1\columnwidth]
        {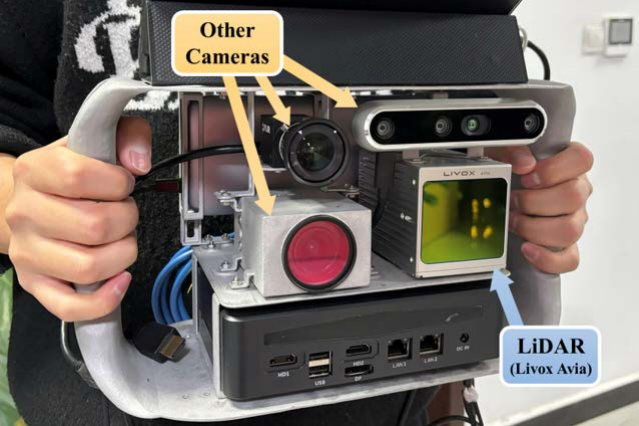}}
    \end{center}
    \vspace{-0.2cm}
    \caption{\label{fig:Sensors}Our data collection device, including the solid-state LiDAR Livox Avia with the built-in BMI088 IMU, and three other cameras.}
    \vspace{0.2cm}
\end{figure}

This paper qualitatively and quantitatively analyzes our AS-LIO framework through a series of experiments. Our sensor suite includes the Livox Avia LiDAR and the built-in BMI088 IMU. To validate our framework, we collected multiple datasets of different scales, environments, and motion conditions, covering indoor and outdoor scenes such as office building corridor, campus road, underground garage, and library. The data collection device is shown in Fig.~\ref{fig:Sensors}.

\subsection{Spatial Overlap Degree Evaluation}

\begin{figure}[t]
    \vspace{1.5mm}
    \begin{center}
        {\includegraphics[width=1\columnwidth]
        {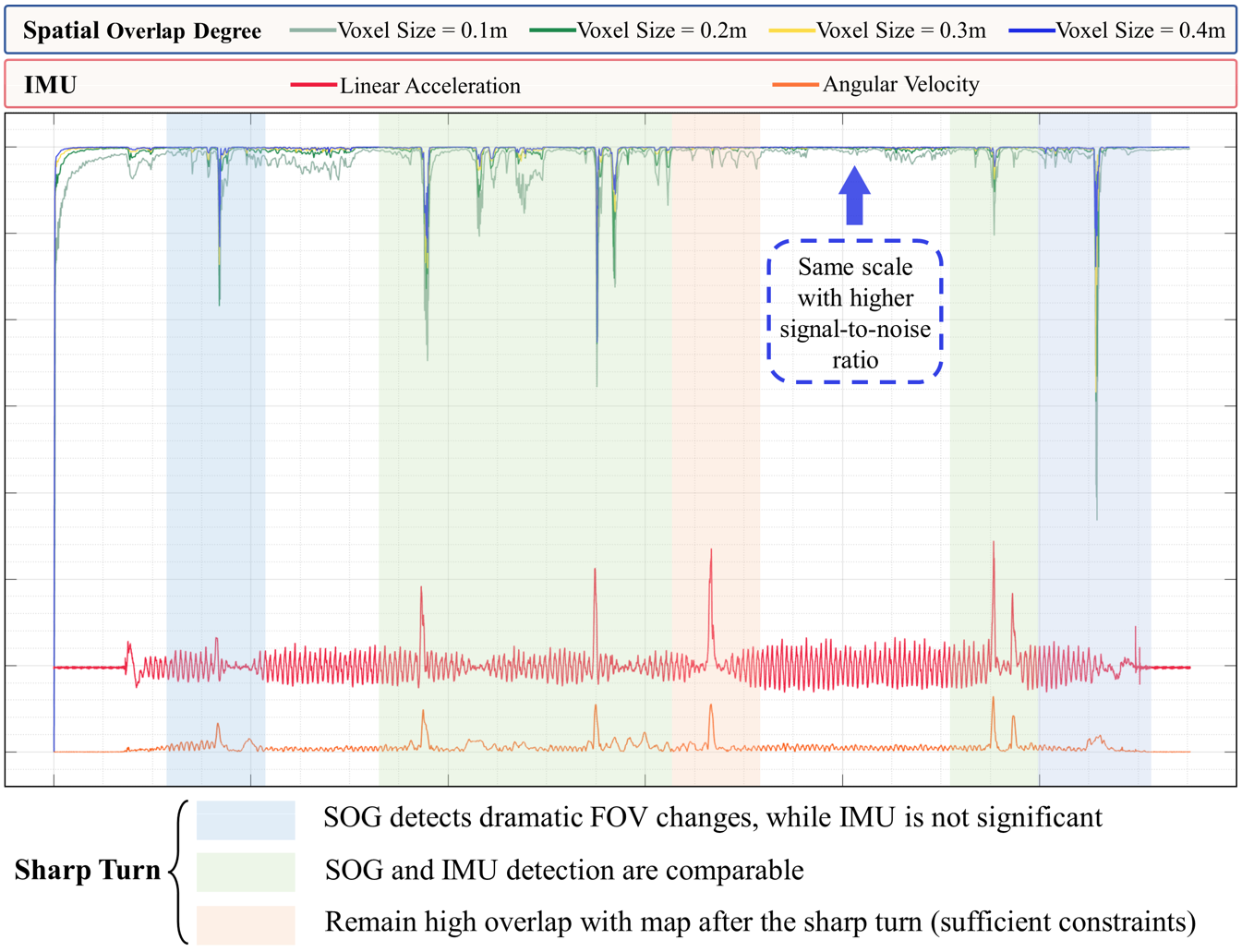}}
    \end{center}
    \vspace{-0.3cm}
    \caption{\label{fig:SOD Evalutation}Spatial overlap degree evalutation. We conduct a sensitivity analysis on the SOD computation, indicating SOD is stable for different voxel size (within reasonable ranges). We also compare IMU data within the same period, indicating SOD with a higher signal-to-noise ratio can significantly identify when FOV change weaken pointcloud constraints.}
    \vspace{-0.3cm}
\end{figure}

We evaluate the SOD with an office building corridor scene dataset \textit{indoor\_1}, and take sharp turns at crossings to enhance the FOV change. The experiment results are shown in Fig.~\ref{fig:SOD Evalutation}. 
Considering that we project the pointcloud frames to the 3d voxel map to calculate the frame-to-map SOD, we conduct sensitivity analysis on the voxel sizes of the map, with ranges of 0.1, 0.2, 0.3, and 0.4 meters. (In general, LIO systems would not adopt too low a map resolution, i.e., too high a voxel size in this case, in order to ensure localization accuracy.)
The experimental results indicate that our SOD exhibits good robustness with respect to the voxel size parameter, and can significantly detect adverse FOV variation with a high signal-to-noise ratio.

\begin{figure}[t]
    \vspace{1.5mm}
    \begin{center}
        {\includegraphics[width=1\columnwidth]
        {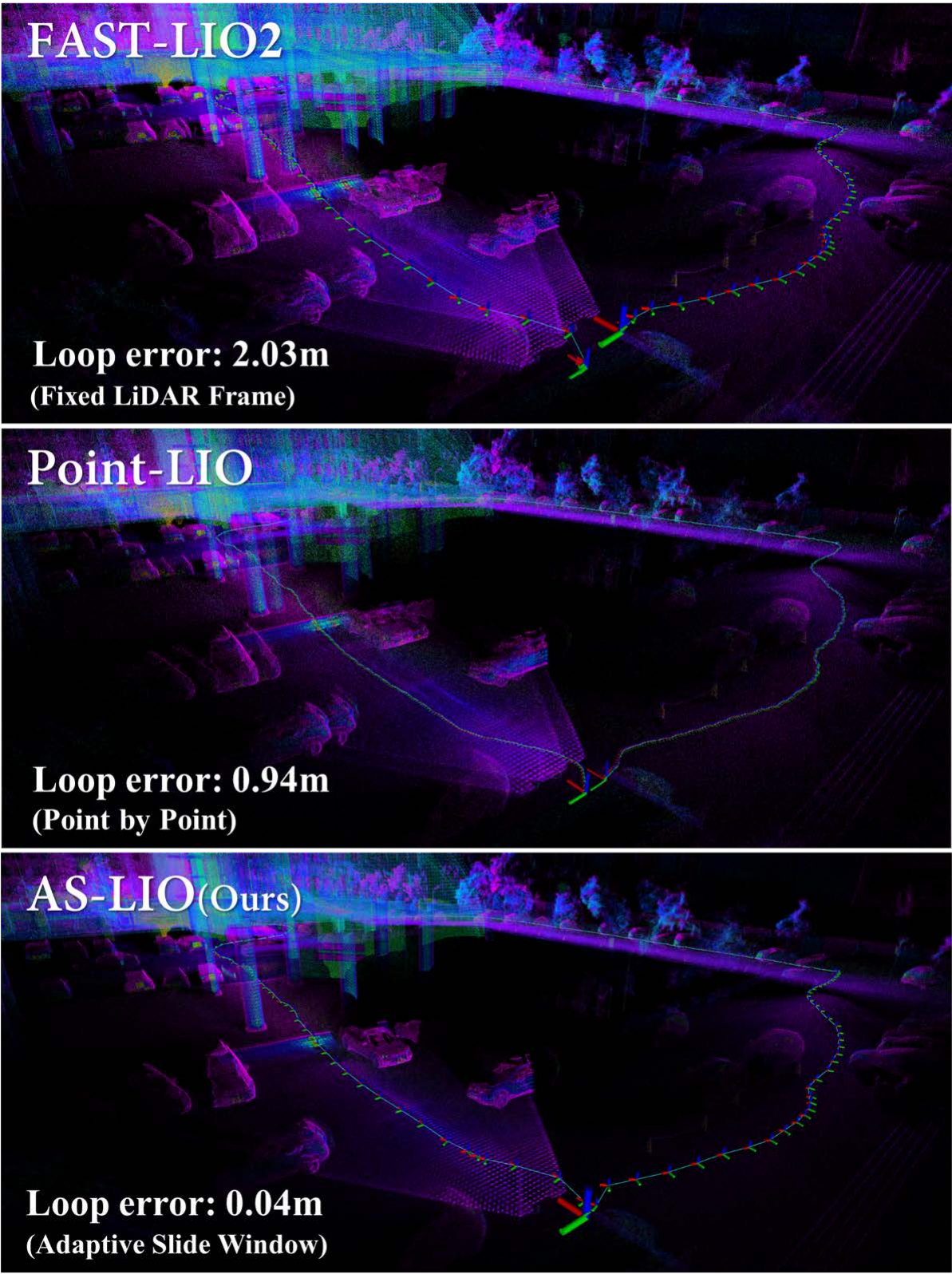}}
    \end{center}
     \vspace{-0.2cm}
    \caption{\label{fig:Accuracy Evaluation}Odometry accuracy evaluation. The figure shows the result of \textit{outdoor\_5}. We complete a strict closed-loop path (about 400m) in campus environment with enhanced sharp turns at crossings. The odometry trajectory and the consistency of map indicate that our AS-LIO achieves better performance.}
     \vspace{-0.3cm}
\end{figure}

In addition, we compared the SOD with the IMU data during corresponding time period.
In turning scenarios, the green area indicates that the detection performance of SOD and IMU is comparable, and the blue area indicates that only SOD is able to significantly detect adverse FOV change. 
In our approach, we aim to measure whether the FOV change reduce the frame-to-map SOD, and weaken the pointcloud constraints thus impacting pointcloud alignment. So that SOD is able to detect the moments when pointcloud constraints are weakened by FOV change, unlike IMU that only detects the motion state.
The red area corresponds to the corridor scene where we turned our heads and returned.
Even though we conducted a rapid turn, the LiDAR's FOV overlapped nicely with the previously established map without much impact on the pointcloud constraints.
We present the SOD and IMU data on the original numerical scale. As shown in the Fig.~\ref{fig:SOD Evalutation}, our SOD has the advantage in terms of signal-to-noise ratio over the IMU noisy raw observations. 
Meanwhile, for scenes that return to the previous FOV, our SOD is able to selectively ignore FOV change with insignificant impact on pointcloud constraints. This allows subsequent refinement steps to be omitted when unnecessary, thus reducing the overall odometry computation cost.

\begin{table}[t]
\footnotesize
\centering
\caption{End to end errors (CM)}
\label{tab:E2E Errors}
\begin{threeparttable}

\begin{tabularx}{0.95\linewidth}{@{}LCCC@{}}
\toprule
 &  {AS-LIO(\footnotesize{Ours})} & {FAST-LIO2} & {Point-LIO}  \\
 \midrule
\begin{tabular}[c]{@{}l@{}} 
\textit{indoor\_1} \\ (\textasciitilde 100m)\end{tabular}   
&  $\textbf{2.25}$ & 28.19 & -\tnote{1}  \\

\begin{tabular}[c]{@{}l@{}} 
\textit{indoor\_2} \\ (\textasciitilde 100m)\end{tabular} 
& 20.88 & $\textbf{14.58}$ & 95.78  \\

\begin{tabular}[c]{@{}l@{}} 
\textit{indoor\_3} \\ (\textasciitilde 100m)\end{tabular} 
& $\textbf{119.88}$ & 247.17 & 249.97 \\
\hline
\begin{tabular}[c]{@{}l@{}} 
\textit{outdoor\_1} \\ (\textasciitilde 300m)\end{tabular}    
& $\textbf{14.38}$  & $\times$\tnote{2}  & 17.24 \\

\begin{tabular}[c]{@{}l@{}} 
\textit{outdoor\_2} \\ (\textasciitilde 300m)\end{tabular} 
& $\textbf{91.64}$ & 227.88  & 230.57 \\

\begin{tabular}[c]{@{}l@{}} 
\textit{outdoor\_3} \\ (\textasciitilde 300m)\end{tabular}   
& $\textbf{112.98}$  & 426.47  &  256.40 \\
\hline
\begin{tabular}[c]{@{}l@{}} 
\textit{outdoor\_4} \\ (\textasciitilde 300m)\end{tabular} 
& $\textbf{96.90}$ & 113.38 & 104.59 \\

\begin{tabular}[c]{@{}l@{}} 
\textit{outdoor\_5} \\ (\textasciitilde 400m)\end{tabular} 
& $\textbf{4.41}$ & 202.51 & 94.37 \\

\begin{tabular}[c]{@{}l@{}} 
\textit{outdoor\_6} \\ (\textasciitilde 500m)\end{tabular}   
& $\textbf{393.31}$  &  483.04  & 410.77 \\

\bottomrule
\end{tabularx}

\begin{tablenotes}
\footnotesize
\item[1] - \,denotes that the system severely diverged midway.
\item[2] × denotes that the system totally failed.
\end{tablenotes}
\end{threeparttable}
    \vspace{-0.4cm}
\end{table}

\subsection{Odometry Accuracy Evaluation}

In this section, we compare the AS-LIO framework mainly with current state-of-the-art LIO systems including FAST-LIO2 and Point-LIO. All methods are implemented in C++ and executed on a laptop equipped with an i7 CPU using the ROS operating system in Ubuntu. These methods only use CPU for computation and do not utilize GPU parallel computing.

Considering that our study aims to improve the accuracy and robustness of LIO system during intense motion, and to rethink how the LIO system can perceive worsened scenarios and diminish errors effectively, we attempted various motion modes during the data collection rather than limiting to smooth, low-speed motion.
To enhance the drasticness of the FOV change, we adopted a more aggressive and rapid turning pattern more aggressive and rapid turning methods at most of corners.

The description of our collected datasets are as follows.

\begin{itemize}

\item  The three indoor dataset scenarios are office building corridors, underground parking garage and library in order. With a restricted FOV in corridor scene, Point-LIO diverged midway in the \textit{indoor\_1} due to the impact of dynamic pedestrians. In the \textit{indoor\_2}, the underground garage is relatively transparent to LiDAR. The columns and vehicles in the scene do not completely blind the LiDAR FOV, and LiDAR can directly scan to the end of the parking garage. While in the \textit{indoor\_3}, the dense bookshelves in the middle of the library scene completely block the light path, and the LiDAR FOV is quite limited.

\item  The first three outdoor datasets were collected in a similar campus road scene but with different motion patterns. 
The \textit{outdoor\_1} was initialized under dynamic conditions, the \textit{outdoor\_2} maintained slight oscillations during walking, and the \textit{outdoor\_3} was entirely collected in running status after initialization.
Lastly, in the last three outdoor datasets (\textit{outdoor\_4, outdoor\_5, outdoor\_6}), we explored more large-scale scenes with sharp turns at each corner to enhance FOV change.

\end{itemize}

Concrete experiment results are shown in Table~\ref{tab:E2E Errors} and Fig.~\ref{fig:Accuracy Evaluation}. The results indicate our AS-LIO framework generally outperforms the FAST-LIO2 and Point-LIO in terms of accuracy, and more effectively mitigating the negative impact of intense motion.

\subsection{Robustness Evaluation in Degradation Scenario}

\begin{figure*}[t]
    \vspace{1.5mm}
    \begin{center}
        {\includegraphics[width=2\columnwidth]
        {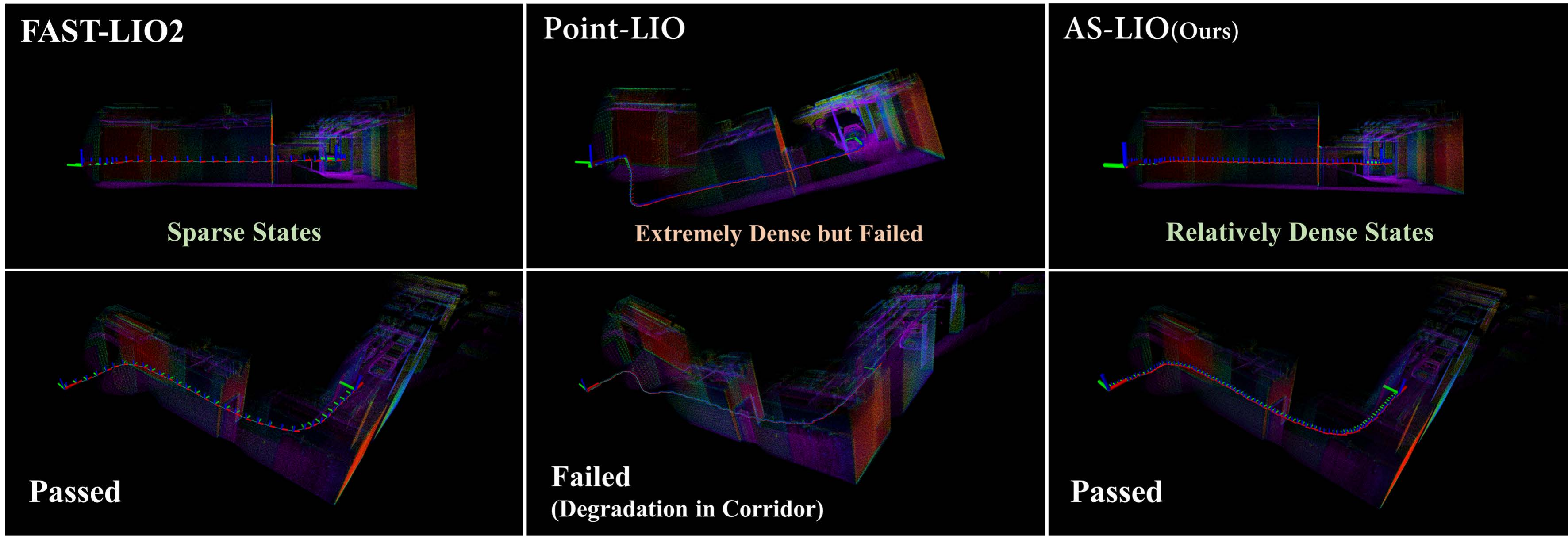}}
    \end{center}
         \vspace{-0.3cm}
    \caption{\label{fig:Robustness Evaluation}Robustness evaluation in underground garage scene. We conduct the test in the scenario with potential degradation risk. The figure shows that AS-LIO ensures the system robustness while maintaining a higher update frequency.}
         \vspace{-0.3cm}
\end{figure*}

To validate the robustness of our AS-LIO framework, we selected scenarios with potential degradation risk to LiDAR for test, and collected a separate section of data in the front corridor of the underground parking garage. 
Concrete test results are shown in Fig.~\ref{fig:Robustness Evaluation}.
The test results indicate that both FAST-LIO2 and our AS-LIO can successfully pass through this scenario, while Point-LIO experiences severe divergence midway due to the instability of constraints.

Our AS-LIO framework does not simply shorten the frame length to refine the trajectory for the sake of accuracy, as the constraint degradation could compromise system stability and introduce unforeseeable risks. 
We adopt the adaptive sliding window that replaces the traditional concept of frame length with an adjustable update step while maintaining historical constraints, thereby reliably enhancing both accuracy in aggressive FOV change. 
Moreover, our sliding window does not maintain additional states, achieving a low computational cost and ensuring the real-time performance.

\addtolength{\textheight}{-7.5cm}
   
\section{CONCLUSIONS}

In this paper we propose AS-LIO, a novel adaptive sliding window LiDAR-Inertial Odometry framework guided by spatial overlap degree.
We rethink how an intelligent LIO system can perceive worsened scenarios and conduct effective error suppression.
Firstly, we evaluate the frame-to-map SOD to directly assess the adverse impact of current FOV change on pointcloud alignment, achieving real-time perception of aggressive FOV variation.
Then, we employ an adaptive sliding window to manage LiDAR stream and control state updates, dynamically adjusting the update step according to the SOD.
In summary, we utilize more dense states to refine trajectory characterizations when necessary, and maintain historical pointcloud constraints within the slide window to enhance data association.

We conduct experiments with a series of datasets containing scenarios at different scales, both indoor and outdoor, and with multiple challenging motion patterns.
The results show that our AS-LIO framework can quickly perceive and respond to the current FOV change, and generally overperforms other benchmark LIO frameworks in terms of accuracy and robustness, suggesting that AS-LIO exhibits better adaptability towards aggressive FOV variation.




\end{document}